# Multi-objective beetle antennae search algorithm

Junfei Zhang[1], Yimiao Huang[1*], Guowei Ma[1,2], Brett Nener[3]

[1]Department of Civil, Environmental and Mining Engineering, The University of Western Australia, Perth 6009, Australia

[2]School of Civil and Transportation Engineering, Hebei University of Technology, 5340 Xiping Road, Beichen District, Tianjin 300401, China

[3]Department of Electrical, Electronic and Computer Engineering, The University of Western Australia, Perth 6009, Australia

* **Corresponding author:** Yimiao Huang

Email: yimiao.huang@uwa.edu.au

**Abstract**

In engineering optimization problems, multiple objectives with a large number of variables under highly nonlinear constraints are usually required to be simultaneously optimized. Significant computing effort are required to find the Pareto front of a nonlinear multi-objective optimization problem. Swarm intelligence based metaheuristic algorithms have been successfully applied to solve multi-objective optimization problems. Recently, an individual intelligence based algorithm called beetle antennae search algorithm was proposed. This algorithm was proved to be more computationally efficient. Therefore, we extended this algorithm to solve multi-objective optimization problems. The proposed multi-objective beetle antennae search algorithm is tested using four well-selected benchmark functions and its performance is compared with other multi-objective optimization algorithms. The results show that the proposed multi-objective beetle antennae search algorithm has higher computational efficiency with satisfactory accuracy.

**Keywords**: Beetle antennae search algorithm; Metaheuristic; Multi-objective optimization; individual intelligence; Engineering design

# 1. Introduction

In engineering optimization problems, multiple objectives that conflict each other are often designed. However, it is very difficult (or impossible) to optimize all the conflicting objectives under highly nonlinear constraints, limited by material properties, design codes and costs (Deb, 2001; Farina, Deb, & Amato, 2004). Algorithms that can be used to solve single-objective optimization problems usually do not work for multi-objective optimization problems. There is no single best solution, but a set of non-dominated solutions should be obtained to approximate the true Pareto front for the multi-objective optimization problem. Usually, the found Pareto solutions do not uniformly distribute along the Pareto front (Gong, Cai, & Zhu, 2009; Yang, 2013). In addition, noises or uncertainties usually exist in real-world optimization problems. A robust optimization algorithm should have high noise tolerance level and produce non-dominated solutions that can sample the search space efficiently. Designers can then select suitable solutions to meet engineering requirements.

Metaheuristic algorithms have been successfully used in solving multi-objective optimization problems (Yang, 2014). Metaheuristic algorithms enhance the efficiency of heuristic procedures by adopting a set of intelligent strategies (usually mimicking successful characteristics in nature (Laporte & Osman, 1995). The widely used multi-objective optimization algorithms include Multi-objective particle swarm optimization (MOPSO)(Reyes-Sierra & Coello, 2006), multi-objective differential evolution (MODE) (Robič & Filipič, 2005), and multi-objective genetic algorithm (NSGA-II) (Konak, Coit, & Smith, 2006). However, all these algorithms are proposed based on swarm-intelligence (SI). For example, the genetic algorithm relies on biological operators (mutation, crossover and selection) to evolve high-quality solutions from a population of candidate solutions. Particle Swarm Optimization simulates the foraging behaviors of birds.

However, for complex engineering problems, a large amount of time is usually required because of the stochastic characteristics of searching approaches of SI algorithms (Kennedy, 2006). Therefore, it is of vital importance to develop robust and efficient metaheuristic algorithms that can obtain optimal solutions under the conditions of limited money, time and resources for real-world optimization problems. Recently, a new metaheuristic algorithm called beetle antennae search (BAS) algorithm has been proposed based on individual intelligence (Jiang & Li, 2017). This algorithm just uses an individual (a beetle) rather than a swarm to search and hence, the calculation time is significantly reduced. Furthermore, it is easy to implement with simple code and less likely to be trapped into local optima by using specific step size strategy (Zhu et al., 2018). These advantages make BAS popular in solving engineering problems such as optimization of hyperparameters of machine learning algorithms for predicting properties of cementitious materials (Sun, Zhang, Li, Ma, et al., 2019; Sun, Zhang, Li, Wang, et al., 2019), global path planning of unmanned aerial vehicles (Wu et al., 2019), and path planning of mobile robots (Wu et al., 2020).

Inspired by the previous successful application of BAS in solving complex engineering problems, this paper extends the BAS algorithm to solve multi-objective optimization engineering problems and develops multi-objective beetle antennae search (MOBAS) algorithm. The MOBAS will be firstly tested validated against a set of benchmark functions and then it will be compared with widely used MOGA, MOEA and MOPSO algorithms. Finally, the advantages and disadvantages of MOBAS as well as the future work will be discussed in the conclusion section.

**2 Basic BAS**

The BAS algorithm mimics the beetle's forging behaviour (Jiang & Li, 2017). The beetle searches for food using its two antennae. When the concentration of odour on the left-antennae side is higher, the beetle moves to the left; otherwise it moves to the right, as shown in Fig. 1. The beetle is simplified to develop the algorithm as shown in Fig. 2. In this model, $x_l$ and $x_r$ represent a position on left-antennae side and right-antennae side, respectively; $x^i$ denotes the position of the beetle at $i^{th}$ time instant ($t$=1, 2…); and $d$ is the distance between the two antennae.

The beetle searches for food in random direction and hence we define a random vector as

$$\mathbf{b} = \frac{rand\ (k, 1)}{\|rand\ (k, 1)\|} \quad (1)$$

where *rand* is a random function and *k* denotes the dimension of the searching space. The position vector of the antennae top can then be written as

$$\mathbf{x}_r^i = \mathbf{x}^i + d^i \mathbf{b} \quad (2)$$

$$\mathbf{x}_l^i = \mathbf{x}^i - d^i\mathbf{b} \tag{3}$$

The position vector of the beetle can be formulated using the following iterative equation:

$$\mathbf{x}^i = \mathbf{x}^{i-1} + \delta^i \mathbf{b}\, sign\left(f(\mathbf{x}_r^i) - f(\mathbf{x}_l^i)\right) \tag{4}$$

where $\delta$ is the step size of the beetle. To avoid local optima, the following step size and antennae length updating strategy can be used:

$$d^i = 0.95^{i-1} + 0.01 \tag{5}$$

$$\delta^i = \delta^{i-1} \tag{6}$$

The pseudocode of BAS is shown in Fig. 3.

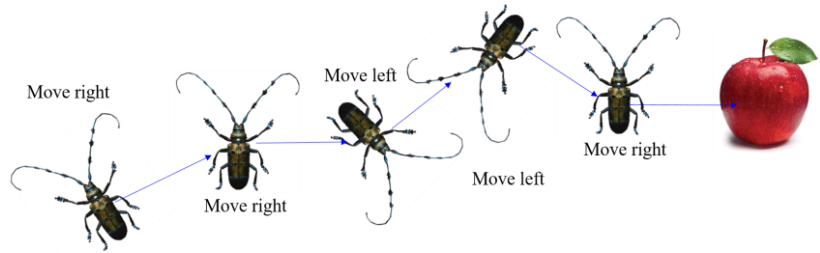

Fig. 1 The foraging behaviour of the beetle.

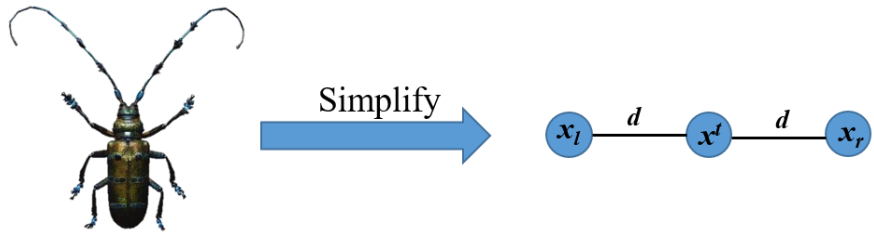

Fig. 2 Simplified beetle model.

**Input** Fitness function $f(\mathbf{x})$, initial position of the beetle $\mathbf{x}^0$, initial step size $\delta^0$, maximum iterations $n$, ratio of antennae length to step size $c$
**Output**: Optimal position $\mathbf{x}_b$, optimal fitness function value $f_b$.
**FOR** $i = 1$ to $n$
    Generate random antennae direction $\mathbf{b}$;
    Calculate the antennae length $d^t = c \times \delta^i$;
    Calculate the left-hand and right-hand positions $\mathbf{x}_l^i$ and $\mathbf{x}_r^i$, respectively;
    Calculate the fitness function value $f(\mathbf{x}_l^i)$ and $f(\mathbf{x}_r^i)$, at the left and right antennae position;
    Calculate the next position $\mathbf{x}^{i+1}$;
    Calculate the fitness function value $f(\mathbf{x}^{i+1})$ at next position $\mathbf{x}^{i+1}$;
    **IF** $f(\mathbf{x}^{i+1}) < f_b$ **THEN**
        Update $\mathbf{x}_b$ to $\mathbf{x}^{i+1}$;
        Update $f_b$ to $f(\mathbf{x}^{i+1})$;
    **END**
    Update step size $\delta^{i+1}$ according to Eq. (4)
    $i = i + 1$;
**END**

Fig. 3 Pseudocode of BAS.

## 3. Multi-objective beetle antennae search algorithm

### 3.1 Multi-objective problem

It is understood multi-objective optimization problems are more computationally intensive to solve than the single one is. In this study, MOBAS is proposed to solve multi-objective optimization problems in engineering.

Several objectives are minimized simultaneously in multi-objective optimization problems. Therefore, the optimality fronts should be found or approximated, which can be written as

$$(\mathbf{x}^*) = \underset{(\mathbf{x})}{\operatorname{argmin}} F(\mathbf{x})$$
$$= \underset{(\mathbf{x})}{\operatorname{argmin}} f_k(\mathbf{x}), x = (1,2,\dots,K) \quad (7)$$

s.t. $\quad g_u(\mathbf{x}) \leq 0, u = \{1,2,\dots\}$

$\quad h_v(\mathbf{x}) = 0, v = \{1,2,\dots\}$

where $f_k(\mathbf{x})$ is the objective function of the decision variable $\mathbf{x}^*$; $g_u(\mathbf{x})$ and $h_v(\mathbf{x})$ specify equality and inequality constraints, respectively. Single-objective optimization is a special case of a multi-objective function with only one objective.

In multi-objective optimization problems, multiple objective functions are minimalized simultaneously. Nonetheless, many real-world problems involve conflicting objectives not to be minimized concurrently. Pareto optimality is introduced to address this issue (Miettinen, 2012). If no objective can be improved without sacrificing at least one other objective, the solutions are identified as non-dominated solutions lying on the Pareto fronts. In mathematical terms, a solution $\mathbf{x}_1$ dominates $\mathbf{x}_2$ if and only if

$$f_k(\mathbf{x}_1) \leq f_k(\mathbf{x}_2), \forall k \in \{1,2,\dots,K\} \quad (10)$$

and

$$f_k(\mathbf{x}_1) < f_k(\mathbf{x}_2), \exists k \in \{1,2,\dots,K\} \quad (11)$$

The Pareto optimal solution $(\mathbf{x}^*)$ is obtained if $F(\mathbf{x}^*)$ dominates $F(\mathbf{x})$ for every $\mathbf{x}$. The Pareto front is defined as the set of Pareto optimal outcomes.

### 3.2 Construction of MOLBAS

In this study, the weighted sum method is used to combine all the objectives $f_k$ into a single objective as follows (Yang, 2013):

$$\Phi = \sum_{k=1}^{K} \varpi_k f_k, \quad \sum_{k=1}^{K} \varpi_k = 1 \tag{12}$$

where weights $\varpi_k$ are generated randomly from a uniform distribution [0, 1]. The randomly generated $\varpi_k$ values can ensure diversity of the non-dominated solutions sample along the Pareto front. To implement the MOBAS, the step size of the beetle is firstly initialized before the objective values of beetle positions are compared. A combined best solution can then be obtained by generating random weight vectors. After the iteration ends, the true Pareto front can be approximated with *n* non-dominated solution points. The pseudocode for multi-objective LBAS using the weighted sum method is shown in below **Algorithm 2**.

**Algorithm 2**: MOBAS

**Input**: Fitness function $F = [f_1(\mathbf{x}_j^i), \ldots, f_k(\mathbf{x}_j^i), \ldots, f_K(\mathbf{x}_j^i)]^T$, initial position of the beetle $X^0 = \{\mathbf{x}_1^0, \ldots, \mathbf{x}_j^0, \ldots\}$, initial step size $\delta^0 = \{\delta\_1^0, \ldots, \delta\_j^0, \ldots\}$, maximum iteration number N, ratio of antennae length to step size c and step size attenuation coefficient α.

**Output**: M optimal Pareto positions (non-dominated solutions) $X_{PF} = \{\mathbf{x}_{PF,1}, \ldots, \mathbf{x}_{PF,m}, \ldots, \mathbf{x}_{PF,M}\}$

m=1;

**WHILE** ($m \leq M$)

    Calculate the random weight of each objective $\mathbf{\Omega} = [\varpi_1, \ldots, \varpi_k, \ldots, \varpi_K]$, and normalized with $\varpi_k = \sum_{k=1}^{K} \varpi_k$;

    **FOR** $i = 1$ to $N$

        Generate random antennae direction $\mathbf{b}^i$;

        Calculate the antennae length $d^i = c \times \delta_m^i$;

        Calculate the left-hand and right-hand positions $\mathbf{x}_l^i$ and $\mathbf{x}_r^i$;

        Calculate the weighted sum function value $\Phi(\mathbf{x}_l^i)$ and $\Phi(\mathbf{x}_r^i)$ at the left and right antennae position with $\Phi(x) = \mathbf{\Omega} \cdot F$;

        Calculate the next position $\mathbf{x}^{i+1} = \mathbf{x}^i + \boldsymbol{\delta}^i$;

        Calculate the weighted sum function value $\Phi(\mathbf{x}^{i+1})$ at next position $\mathbf{x}^{i+1}$;

        **IF** $\Phi(\mathbf{x}^{i+1}) < \Phi_b$ **THEN**

            Update $\mathbf{x}_b$ to $\mathbf{x}^{i+1}$;

            Update $\Phi_b$ to $\Phi(\mathbf{x}^{i+1})$;

        **END IF**

        Update step size $\boldsymbol{\delta}^{i+1} = \alpha \boldsymbol{\delta}^i$;

    **END FOR**

    **IF** $\mathbf{x}_b$ satisfy all the constraints

        **IF** $\mathbf{x}_b$ is not dominated by $X_{PF}$, **THEN**

            Update $X_{PF} = X_{PF} \cap x_b$;

            Update $m = m + 1$;

        **END IF**

        **FOR** $X_{PF,t}$ **IN** $X_{PF}$

            **IF** $x_b$ dominates $x_{PF,t}$, **THEN**

                Update $X_{PF} = X_{PF} - x_{PF,t}$;

                Update $m = m - 1$

            **END IF**

        **END FOR**

    **END IF**

**END WHILE**

# 4 Numerical results

## 4.1 Multi-objective test functions

A number of multi-objective benchmark functions have been used to test multi-objective optimization algorithms in literature (Zhang et al., 2008; Zitzler, Deb, & Thiele, 2000; Zitzler & Thiele, 1999). Among these benchmark functions, we selected the following ones with discontinuous, non-convex and convex Pareto fronts to validate the proposed MOBAS.

- Schaffer's Min-Min (SCH) function with convex Pareto front

$$f_1(x) = x^2,$$
$$f_2(x) = (x-2)^2, \quad (13)$$
$$-10^3 \le x \le 10^3$$

The Pareto front for SCH is $f_2 = \left(\sqrt{f_1} - 2\right)^2$.

- ZDT 1 function with convex front

$$f_1(x) = x_1,$$
$$f_2(x) = g(1 - \sqrt{f_1/g}),$$
$$g(x_2, \dots, x_n) = 1 + \frac{9}{n-1}\sum_{i=2}^{n} x_i, \quad (14)$$
$$x_i \in [0,1], i = 1, \dots, n$$

where *n* denotes the dimension number. The Pareto front for ZDT 1 is $f_2 = 1 - \sqrt{f_1}$.

- ZDT 2 function with a non-convex front

$$f_1(x) = x_1,$$
$$f_2(x) = g\left(1 - \frac{f_1}{g}\right)^2,$$
$$g(x_2, \dots, x_n) = 1 + \frac{9}{n-1}\sum_{i=2}^{n} x_i \quad (15)$$
$$x_i \in [0,1], i = 1, \dots, n$$

where *n* denotes the dimension number. The Pareto front for ZDT 2 is $f_2 = 1 - f_1^2$.

- ZDT 3 function with a discontinuous front

$$f_1(x) = x_1, \quad (16)$$

$$f_2(x) = g[1 - \sqrt{\frac{f_1}{g}} - \frac{f_1}{g}\sin(10\pi f_1)],$$

$$g(x_2, \ldots, x_n) = 1 + \frac{9}{n-1}\sum_{i=2}^{n} x_i$$

$$x_i \in [0,1], i = 1, \ldots, n$$

where $n$ denotes the dimension number. The Pareto front for ZDT 3 is $f_2 = 1 - \sqrt{f_1} - f_1\sin(10\pi f_1)$ with $f_1 \in F = [0, 0.083] \cup (0.182, 0.258] \cup (0.409, 0.454] \cup (0.618, 0.653] \cup (0.823, 0.852]$.

We generated 200 Pareto points by MOBAS and compared these points with the true Pareto front (PF) for SCH, ZDT 1, ZDT 2 and ZDT 3, as shown in Fig. 4. It can be seen that all the found Pareto point lie on the true Pareto front without visible errors. To evaluate the errors, we defined the following equations:

$$AD^i = \frac{1}{M}\sqrt{\sum_{m=1}^{M}\left[\Psi\left(f_1(x_m^i)\right) - f_2(x_m^i)\right]^2} \qquad (17)$$

where $AD^i$ is the error or distance between the estimated Pareto point $(f_1(x^i), f_2(x^i))$ at position $x^i$ and actual Pareto front $\Psi$; $M$ denotes the Pareto points found by MOBAS.

Fig. 5 shows the decrease of the error with increasing iteration. It is evident that the MOBAS decreases exponentially for all benchmark functions, indicating that the MOBAS is feasible and has high efficiency.

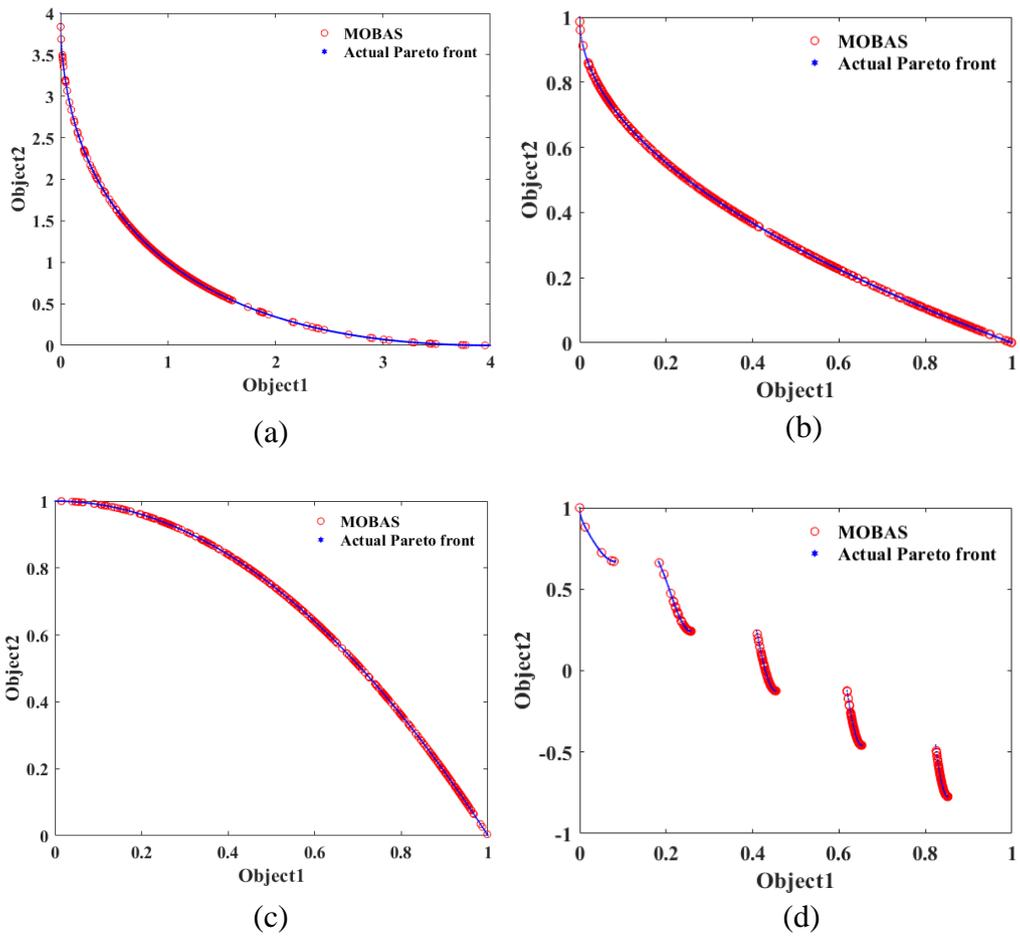

Fig. 4 Pareto front of function SCH (a), ZDT 1 (b), ZDT 2 (c) and ZDT 3 (d).

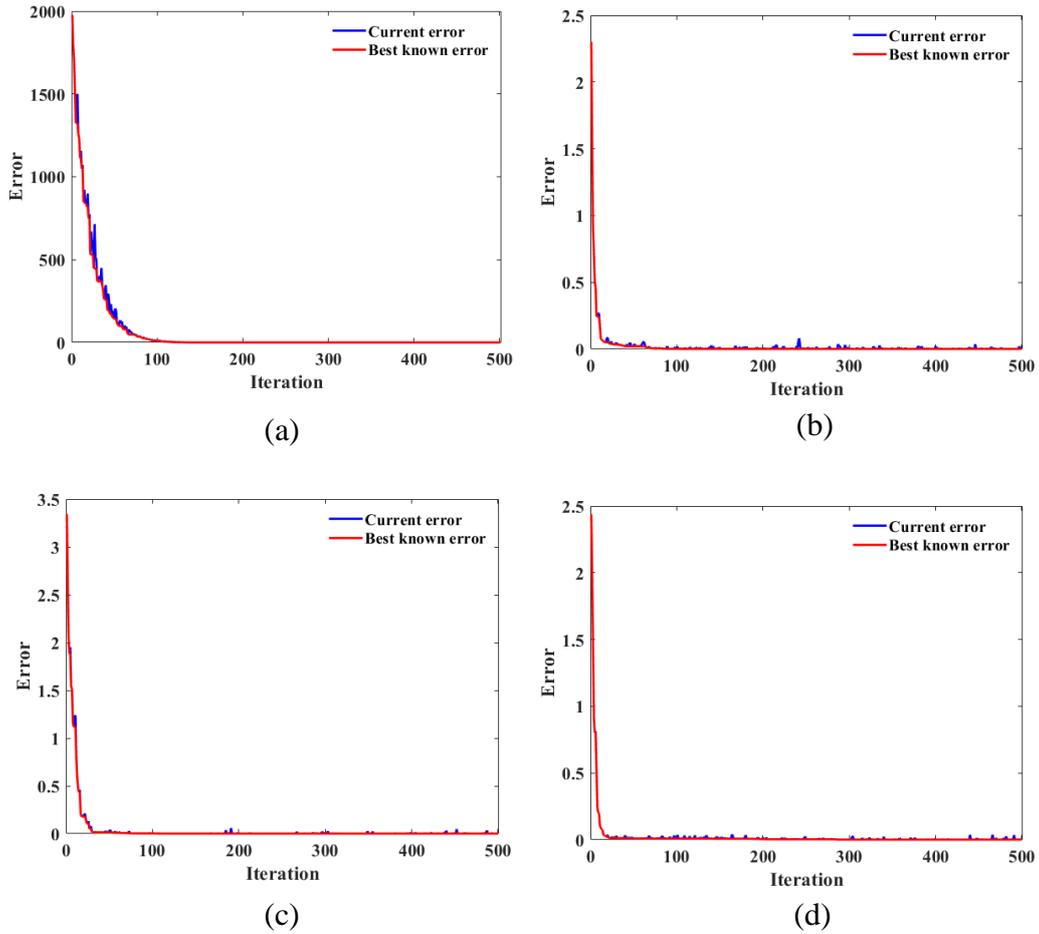

Fig. 5 Convergence of the MOBAS with the first 500 iterations for SCH (a), ZDT 1 (b), ZDT 2 (c) and ZDT 3 (d).

**4.2 Comparative study**

Multi-objective Particle Swarm Optimization (MOPSO) (Mostaghim & Teich, 2003), Non-dominated Sorting Genetic Algorithm II (NSGA-II) (Deb, Pratap, Agarwal, & Meyarivan, 2002), and multi-objective differential evolution (MODE) (Babu & Gujarathil, 2007) have been widely used to solve multi-objective optimization problems. Therefore, this paper compares the proposed MOBAS with the above three multi-objective algorithms. As 200 Pareto points need to be found, the population size of the swarm intelligence algorithm is selected to be 200. For MOPSO, the following parameters are used: inertia weight=0.5, inertia weight damping rate=0.99, personal learning coefficient=1, global learning coefficient=2. The parameters used for NSGA-II are as follows: crossover percentage=0.7, mutation percentage=0.4, mutation rate=0.02, mutation step size= one-tenth of domain length. The MOEA algorithm has the upper and lower limits of neighbours to be 15 and 2, respectively, neighbour number to population size ratio of 0.15 and crossover percentage of 0.5. The results are tabulated in Table 1. It can be seen that the proposed MOBAS obtained better results for SCH function. For ZDT 1, ZDT 2 and ZDT 3 functions the error obtained by MOBAS is better than NSGA-II and MOEA but slightly worse than MOPSO (still the same order). To further improve the accuracy of MOBAS, searching strategies such as Levy flight, self-adaptive inertia weight and chaotic maps can be introduced to avoid premature converge and falling into local minima. It should be noted that for all the four functions, the computing time of MOBAS for obtaining 200 Pareto points is much smaller than that of other multi-objective algorithms, indicating high computational efficiency of MOBAS. This is of vital importance when solving multi-objective engineering problems, as it takes a large amount of time to optimize many objectives of engineering with highly-nonlinear constrains.

Table 1 Comparison of performances of different algorithms.

| Error | SCH | | ZDT 1 | | ZDT 2 | | ZDT 3 | |
|---|---|---|---|---|---|---|---|---|
| | Elapsed time (s) | $AD^M$ | Elapsed time (s) | $AD^M$ | Elapsed time (s) | $AD^M$ | Elapsed time (s) | $AD^M$ |
| **MOBAS** | **35.35** | **3.52E-05** | **76.11** | 1.55E-03 | **77.77** | 9.59E-04 | **147.95** | 7.48E-03 |
| NSGA-II | 449.53 | 5.73E-03 | 442.75 | 3.33E-02 | 448.36 | 7.24E-02 | 456.71 | 1.14E-01 |
| MOEA | 926.62 | 9.32E-04 | 974.01 | 5.80E-03 | 947.11 | 5.50E-03 | 933.36 | 2.15E-02 |
| MOPSO | 137.57 | 1.79E-04 | 232.05 | **1.08E-03** | 228.51 | **7.55E-04** | 243.04 | **1.18E-03** |

## 5. Conclusions

It is very difficult to solve multi-objective engineering optimization problems. A new multi-objective optimization algorithm (MOBAS) was recently formulated in this study based on the recently proposed beetle antennae search algorithm. The proposed MOBAS was tested on four benchmark functions.

The performance of MOBAS was compared with other multi-objective algorithms. The results show that MOBAS is more computationally efficient with satisfactory accuracy. The future work can focus on introducing some premature-preventing strategies such as Levy flight, self-adaptive inertia weight and chaotic maps to further improve the accuracy of MOBAS.